\documentclass[runningheads]{llncs}
\usepackage{graphicx}

\usepackage{tikz}
\usepackage{comment}
\usepackage{amsmath,amssymb} 
\usepackage{color}

\usepackage[accsupp]{axessibility}  

\usepackage{pgfplots}
\DeclareUnicodeCharacter{2212}{−}
\usepgfplotslibrary{groupplots,dateplot}
\usepackage{tikz}
\usetikzlibrary{patterns,shapes.arrows,positioning}
\pgfplotsset{compat=newest}

\usepackage{array}
\usepackage{multirow}
\usepackage{booktabs}
\usepackage{adjustbox}

\usepackage{caption} 
\usepackage{subcaption}

\usepackage{wrapfig}

\makeatletter
\DeclareRobustCommand\onedot{\futurelet\@let@token\@onedot}
\def\@onedot{\ifx\@let@token.\else.\null\fi\xspace}

\makeatother

\newcommand{\oursVGG}{\small Ours (VGG-F)}
\newcommand{\oursFlickr}{\small Ours (Flickr-F)}
\newcommand{\sImagenet}{\small Sup. (ImageNet)}
\newcommand{\sVGG}{\small Sup. (VGG-F)}
\newcommand{\scratch}{\small Scratch}

\DeclareMathOperator*{\argmin}{argmin}

\begin{document}
\pagestyle{headings}
\mainmatter

\title{Pre-training strategies and datasets for facial representation learning} 

\titlerunning{Pre-training strategies and datasets for facial representation learning}
%
\author{Adrian Bulat\inst{1}\orcidID{0000-0002-3185-4979} \and
Shiyang Cheng\inst{1}, Jing Yang\inst{2}\orcidID{0000-0002-8794-4842}, Andrew Garbett\inst{1}, Enrique Sanchez\inst{1}\orcidID{0000-0003-0196-922X} \and Georgios Tzimiropoulos\inst{1,3}\orcidID{0000-0002-1803-5338} 
}
\authorrunning{A. Bulat et al.}
%
\institute{Samsung AI Cambridge \\
\email{adrian@adrianbulat.com}, \email{\{shiyang.c,a.garbett\}@samsung.com}, \email{kike.sanc@gmail.com}\\
\and
University of Nottingham, Nottingham, UK \\
\email{jing.yang2@nottingham.ac.uk}
\and
Queen Mary University London, London, UK \\
\email{g.tzimiropoulos@qmul.ac.uk}}
\maketitle

\begin{abstract}
 What is the best way to learn a universal face representation? Recent work on Deep Learning in the area of face analysis has focused on supervised learning for specific tasks of interest (e.g. face recognition, facial landmark localization etc.) but has overlooked the overarching question of how to find a facial representation that can be readily adapted to several facial analysis tasks and datasets. To this end, we make the following 4 contributions: (a) we introduce, for the first time, a comprehensive evaluation benchmark for facial representation learning consisting of 5 important face analysis tasks. (b) We systematically investigate two ways of large-scale representation learning applied to faces: supervised and unsupervised pre-training. Importantly, we focus our evaluations on the case of few-shot facial learning. (c) We investigate important properties of the training datasets including their size and quality (labelled, unlabelled or even uncurated). (d) To draw our conclusions, we conducted a very large number of experiments. Our main two findings are: (1) Unsupervised pre-training on completely in-the-wild, uncurated data provides consistent and, in some cases, significant accuracy improvements for all facial tasks considered. (2) Many existing facial video datasets seem to have a large amount of redundancy. We will release code, and pre-trained models to facilitate future research. 
\end{abstract}

\section{Introduction}

Supervised learning with Deep Neural Networks has been the standard approach to solving several Computer Vision problems over the recent past years~\cite{he2016deep,ren2015faster,simonyan2014very,huang2017densely,long2015fully}. Among others, this approach has been very successfully applied to several face analysis tasks including face detection~\cite{cai2016unified,zhang2017s3fd,deng2019retinaface,li2019dsfd}, recognition~\cite{schroff2015facenet,wang2018additive,wang2018cosface,yang2020fan,deng2019arcface} and landmark localization~\cite{bulat2016two,bulat2017far,zhu2017face,wang2020deep}. For example, face recognition was one of the domains where even very early attempts in the area of deep learning demonstrated performance of super-human accuracy~\cite{phillips2018face,taigman2014deepface}. Beyond deep learning, this success can be largely attributed to the fact that for most face-related application domains, large scale datasets could be readily collected and annotated, see for example~\cite{cao2018vggface2,bulat2017far}. 

There are several concerns related to the above approach. Firstly, from a practical perspective, collecting and annotating new large scale face datasets is still necessary; examples of this are context-dependent domains like emotion recognition~\cite{gunes2010automatic,tzirakis2017end,valstar2016avec} and surveillance~\cite{burton1999face,grgic2011scface}, or new considerations of existing problems like fair face recognition~\cite{robinson2020face,sixta2020fairface}. Secondly, from a methodological point of view, it is unsatisfactory for each application to require its own large-scale dataset, although there is only one object of interest - the human face. 

To this end, we investigate, for the first time to our knowledge, the task of large-scale learning universal facial representation in a principled and systematic manner. In particular, we shed light to the following research questions:
\begin{itemize}
\item
\textit{``What is the best way to learn a universal facial representation that can be readily adapted to new tasks and datasets? Which facial representation is more amenable to few-shot facial learning?''}
\item
\textit{``What is the importance of different training dataset properties (including size and quality) in learning this representation? Can we learn powerful facial feature representations from uncurated facial data as well?''}
\end{itemize}
\begin{figure*}[t]
    \centering
    \includegraphics[width=11cm, trim=0 9cm 0 3cm, clip]{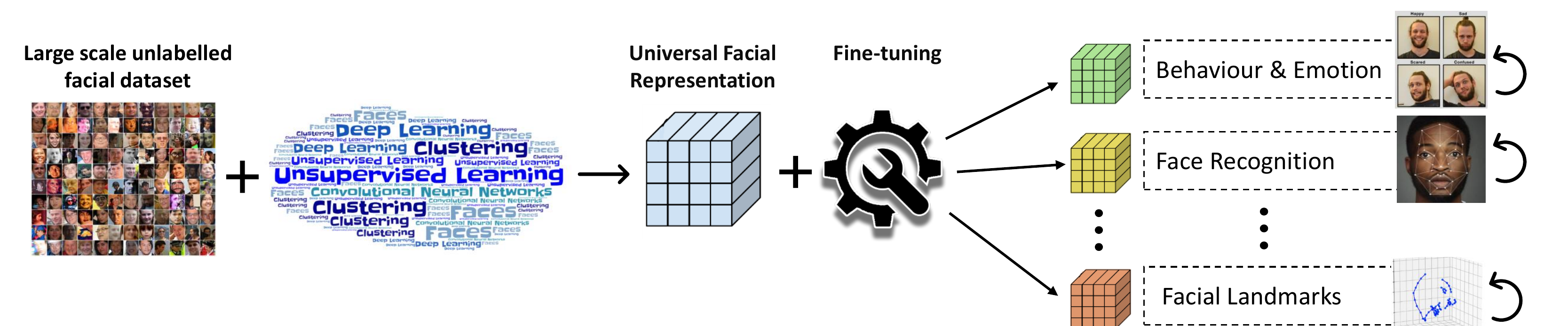}
    \caption{We advocate for a new paradigm to solving face analysis based on the following pipeline: (1) collection of large-scale unlabelled facial dataset, (2) (task agnostic) network pre-training for universal facial representation learning, and (3) facial task-specific fine-tuning. \textbf{Our main result} is that even when training on a completely in-the-wild, uncurated dataset downloaded from Flickr, this generic pipeline provides consistent and, in some cases, significant accuracy improvements for all facial tasks considered.}
    \label{fig:overall}
\end{figure*}

To address this, \textbf{we make} the following \textbf{4 contributions}:
\begin{enumerate}
\item
We introduce, for the first time, a comprehensive and principled evaluation benchmark for facial representation learning consisting of 5 important face analysis tasks, namely face recognition, AU recognition, emotion recognition, landmark localization and 3D reconstruction. 
\item
Within this benchmark, and for the first time, we systematically evaluate 2 ways of large-scale representation learning applied to faces: supervised and unsupervised pre-training. Importantly, we focus our evaluations on the case of few-shot facial learning where only a limited amount of data is available for the downstream tasks.
\item
We systematically evaluate the role of datasets in learning the facial feature presentations by constructing training datasets of varying size and quality. To this end, we considered ImageNet, several existing curated face datasets but also a new in-the-wild, uncurated face dataset downloaded from Flickr.  
\item
We conducted extensive experiments to answer the aforementioned research questions and from them we were able to draw several interesting observations and conclusions. 

\end{enumerate}

Our main findings are: (a) Even when training on a completely in-the-wild, uncurated dataset downloaded from Flickr, unsupervised pre-training pipeline provides consistent and, in some cases, significant accuracy improvements for all facial tasks considered. (b) We found that many existing facial video datasets seem to have a large amount of redundancy. Given that unsupervised pre-training is cheap and that the cost of annotating facial datasets is often significant, some of our findings could be particularly important for researchers when collecting new facial datasets is under consideration. Finally, we will release code and pre-trained models to facilitate future research.  

\section{Related Work}

\noindent \textbf{Facial transfer learning:} Transfer learning in Computer Vision typically consists of ImageNet pre-training followed by fine-tuning on the downstream task~\cite{ren2015faster,dai2016r,chen2014semantic}. Because most recent face-related works are based on the collection of larger and larger facial datasets~\cite{guo2016ms,bulat2017far,mollahosseini2017affectnet}, the importance of transfer learning has been overlooked in face analysis and, especially, the face recognition literature. ImageNet pre-training has been applied to face analysis when training on small datasets is required, for example for emotion recognition~\cite{ng2015deep}, face anti-spoofing ~\cite{parkin2019recognizing} and facial landmark localization~\cite{wang2020deep}. Furthermore, the VGG-Face~\cite{parkhi2015deep} or other large face datasets (e.g.~\cite{mollahosseini2017affectnet}) have been identified as better alternatives by several works, see for example~\cite{fan2016video,kaya2017video,yang2017neural,knyazev2017convolutional,ranjan2017all,ranjan2017hyperface,parkin2019recognizing,kossaifi2020factorized,li2020deep}. To our knowledge, we are the first to systematically evaluate supervised network pre-training using both ImageNet and VGG-Face datasets on several face analysis tasks.

\noindent \textbf{Facial datasets:} The general trend is to collect larger and larger facial datasets for the face-related task in hand~\cite{guo2016ms,bulat2017far,mollahosseini2017affectnet}. Also it is known that label noise can severely impact accuracy (e.g. see Table 6 of~\cite{deng2019arcface}). Beyond faces, the work of~\cite{mahajan2018exploring} presents a study which shows the benefit of \textit{weakly supervised} pre-training on much larger datasets for general image classification and object detection. Similarly, we also investigate the impact of the size of facial datasets on \textit{unsupervised pre-training} for facial representation learning. Furthermore, one of our main results is to show that a high-quality facial representation can be learned even when a completely uncurated face dataset is used.

\noindent \textbf{Few-shot face analysis:} Few-shot  refers to both low data and label regime. There is very little work in this area. To our knowledge, there is no prior work on few-shot face recognition where the trend is to collect large-scale datasets with millions of samples (e.g.~\cite{guo2016ms}). There is no systematic study for the task of emotion recognition, too. There is only one work on few-shot 
learning for facial landmark localization, namely that of~\cite{browatzki20203fabrec} which, different to our approach, proposes an auto-encoder approach for network pre-training. To our knowledge, our evaluation framework provides the very first comprehensive attempt to evaluate the transferability of facial representations for few-shot learning for several face analysis tasks. 

\noindent \textbf{Semi-supervised face analysis:}  Semi-supervised learning has been applied to the domain of Action Unit recognition where data labelling is extremely laborious~\cite{zhang2018weakly,zhang2019joint,zhang2019context}. Although these methods work with few labels, they are domain specific (as opposed to our work), assuming also that extra annotations are available in terms of ``peak" and ``valley" frames which is also an expensive operation.

\noindent \textbf{Unsupervised learning:} There is a very large number of recently proposed unsupervised/self-supervised learning methods, see for example~\cite{wu2018unsupervised,caron2018deep,ye2019unsupervised,misra2019self,he2019momentum,chen2020simple,caron2020unsupervised,chen2020big,grill2020bootstrap}. To our knowledge, only very few attempts from this line of research have been applied to faces so far. The authors of~\cite{wiles2018self} learn face embeddings in a self-supervised manner by predicting the motion field between two facial images. The authors of~\cite{vielzeuf2019towards} propose to combine several facial representations learned using an autoencoding framework. In this work, we explore learning facial representations in an unsupervised manner using the state-of-the-art method of~\cite{caron2020unsupervised} and show how to effectively fine-tune the learned representations to the various face analysis tasks of our benchmark.

\section{Method}

Supervised deep learning directly applied to large labelled datasets is the de facto approach to solving the most important face analysis tasks. In this section, we propose to take a different path to solving face analysis based on the following 2-stage pipeline: (task agnostic) network pre-training followed by task adaptation. Importantly, we argue that network pre-training should be actually considered as part of the method and not just a simple initialization step. We explore two important aspects of network pre-training: (1) the method used, and (2) the dataset used. Likewise, we highlight hyper-parameter optimization for task adaptation as an absolutely crucial component of the proposed pipeline. Finally, we emphasize the importance of evaluating face analysis on low data regimes, too. We describe important aspects of the pipeline in the following sections.

\subsection{Network Pre-training}

\noindent \textbf{Supervised pre-training} of face networks on ImageNet or VGG datasets is not new. We use these networks as strong baselines. For the first time, we comprehensively evaluate their impact on the most important face analysis tasks.

\noindent \textbf{Unsupervised pre-training:} Inspired by~\cite{grill2020bootstrap,misra2019self,he2020momentum,caron2020unsupervised}, we explore, for the first time in literature, large-scale unsupervised learning on facial images to learn a universal, task-agnostic facial representation. To this end, we adopt the recently proposed SwAV~\cite{caron2020unsupervised} which simultaneously clusters the data while enforcing consistency between the cluster assignments produced for different augmentations of the same image. The pretext task is defined as a ``swapped'' prediction problem where the code of one view is predicted from the representation of another:
$\mathcal{L}(\mathbf{z}_{0}, \mathbf{z}_{1}) = \ell(\mathbf{z}_{0}, \mathbf{q}_{1}) + \ell(\mathbf{z}_{1}, \mathbf{q}_{0}),$
where $\mathbf{z}_{0}, \mathbf{z}_{1}$ are the features produced by the network for two different views of the same image and $\mathbf{q}_{0}, \mathbf{q}_{1}$ their corresponding codes computed by matching these feature using a set of prototypes. $\ell$ is a cross-entropy (with temperature) loss. 
See supplementary material for training details.

\subsection{Pre-training Datasets}\label{sec:method-datasets}

With pre-training being now an important part of the face analysis pipeline, it is important to investigate what datasets can be used to this end. We argue that supervised pre-training is sub-optimal due to two main reasons: (a) the resulting models may be overly specialized to the source domain and task (e.g. face recognition pre-training) or be too generic (e.g. ImageNet pre-training), and (b) the amount of labeled data may be limited and/or certain parts of the natural data distribution may not be covered. 
To alleviate this, for the first time, we propose to explore large scale unsupervised pre-training on 4 facial datasets of interest, under two settings: using curated and uncurated data. The later departs from the common paradigm that uses carefully collected data that already includes some forms of explicit annotations and post-processing. In contrast, in the later case, all acquired facial images are used.
\subsubsection{Curated Datasets}

For unsupervised pre-training we explore 3 curated datasets, collected for various facial analysis tasks:  (a) Full VGG-Face ($\sim 3.4M$), (b) Small VGG-Face ($\sim1M$) and (c) Large-Scale-Face ($ > 5.0M$), consisting of VGG-Face2~\cite{cao2018vggface2}, 300W-LP~\cite{zhu2016face}, IMDb-face~\cite{wang2018devil}, AffectNet~\cite{mollahosseini2017affectnet} and WiderFace~\cite{yang2016wider}. During unsupervised pre-training we drop all labels using only the facial images. See supplementary material for more details.

\subsubsection{Uncurated Datasets}

For a more realistic and practical scenario, we go beyond sanitized datasets, by creating a completely uncurated, in-the-wild, dataset, coined Flickr-Face, of $\sim1.5M$ facial images by simply downloading images from Flickr (using standard search keywords like ``faces'', ``humans'', etc.) and filtering them with a face detector~\cite{deng2019retinaface} (the dataset will be made available). In total we collected 1.793.119 facial images. For more details, see supp. material.

\subsection{Facial Task Adaptation}
\noindent \textbf{End facial tasks:} To draw as safe conclusions as possible, we used a large variety of face tasks (5 in total) including face recognition (classification), facial Action Unit intensity estimation (regression), emotion recognition in terms of valence and arousal (regression), 2D facial landmark localization (pixel-wise regression), and 3D face reconstruction (GCN regression). For these tasks, we used, in total, 10 datasets for evaluation purposes.

\captionsetup{skip=1pt}
\setlength\intextsep{0pt}
\begin{wrapfigure}[25]{r}{0.5\textwidth}
    \centering
    \includegraphics[width=6.0cm]{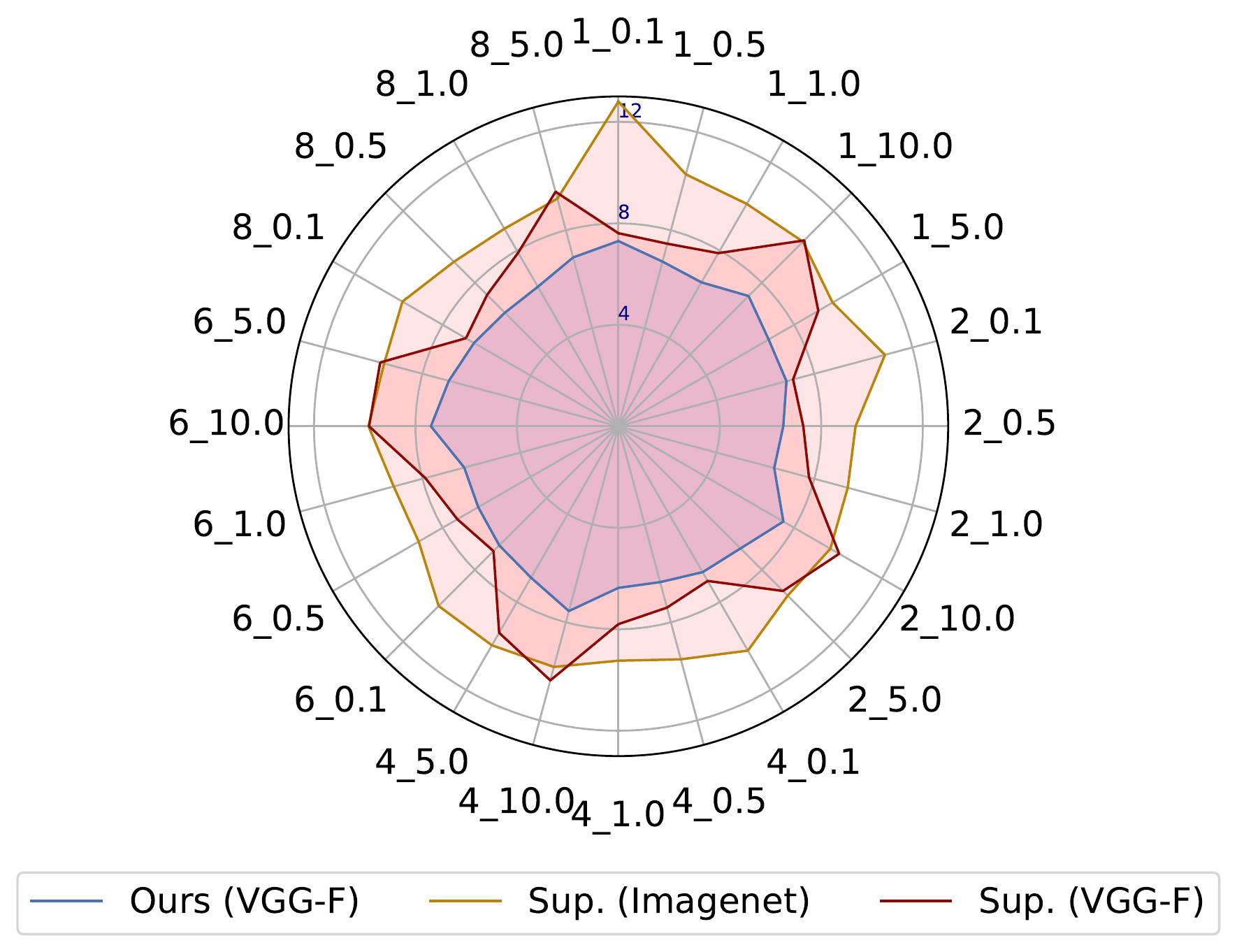}
    \caption{Facial landmark localization accuracy in terms of NME (\%) of 3 different pre-training methods for selected combinations of hyperparameters. The labels on the figure's perimeter show the scheduler length (first value) and backbone relative's learning rate (second value) separated by an underscore. Each circle on the radar plot denotes a constant error level. Points located closer to the center correspond to lower error levels. Accuracy greatly varies for different hyperparameters.}
    \label{fig:hyperprams-fa}
\end{wrapfigure}

\noindent \textbf{Adaptation methods:} We are given a pre-trained model on task $m$, composed of a backbone $g(.)$ and a network head $h^{m}(.)$. The model follows the ResNet-50~\cite{he2016deep} architecture. We considered two widely-used methods for task adaptation: 
(a) \textit {Network fine-tuning} adapts the weights of $g(.)$ to the new task $m_i$. The previous head is replaced with a task-specific head $h^{m_i}(.)$ that is trained from scratch. (b) \textit{Linear layer adaptation} keeps the weights of $g(.)$ fixed and trains only the new head $h^{m_i}(.)$. Depending on the task, the structure of the head varies. This will be defined for each task in the corresponding section. See also Section~\ref{sec:results}.

\noindent \textbf{Hyper-parameter optimization:} We find that, without a proper hyper-parameters selection for each task and setting, the produced results are often misleading. In order to alleviate this and ensure a fair comparison, we search for the following optimal hyper-parameters: (a) learning rate, (b) scheduler duration and (c) backbone learning rate for the pre-trained ResNet-50. This search is repeated for \textit{each data point} defined by the tuple (task, dataset, pre-training method and $\%$ of training data). In total, this yields in an extraordinary number of experiments for discovering the optimal hyperparameters.

Fig.~\ref{fig:hyperprams-fa} shows the importance of hyperparameters on accuracy for the task of facial landmark localization. In particular, for 1 specific value of learning rate, about 40 different combinations of scheduler duration and backbone relative's learning rate are evaluated. 24 of those combinations are placed on the perimeter of the figure. The 3 closed curves represent the Normalized Mean Error (NME) for each hyperparameter combination for each pre-training method. We observe that accuracy greatly varies for different hyperparameters.

\begin{wraptable}[12]{r}{0.4\textwidth}
\caption{Comparison between the facial representations learned by MoCov2 and SwAV, by fine-tuning the models on 2\% of 300W and DISFA.} 
\label{tab:results_moco_vs_swav}
\begin{center}
\resizebox{.38\textwidth}{!}{
    \begin{tabular}{lcc}
    \toprule
    \multirow{2}{1.1cm}{\textbf{Method}} & 
     \textbf{300W} & \textbf{DISFA}
     \\
     \cmidrule(lr){2-3}
     & NME (\%) & ICC   \\
    \midrule
    Scratch & 13.5 &  .237 \\
    MoCov2 & 11.9 &  .280 \\
    SwAV & \textbf{4.97} & \textbf{.560} \\
    SwAV (256) & 5.00 & .549 \\
    \bottomrule
    \end{tabular}
    }
\end{center}
\end{wraptable}

\newpage
\subsection{Few-shot Learning-based Evaluation}\label{ssec:method-fewshot}
We explore, for the first time, evaluating the face models using a varying percentage of training data for each face analysis task. Specifically, beyond the standard evaluation using $100\%$ of training data, we emphasize the importance of the low data regime, in particular $10\%$ and $2\%$, which has a clear impact when new datasets are to be collected and annotated. The purpose of the proposed evaluation is not only to show which method works the best for this setting but also to draw interesting conclusions about the redundancy of existing facial datasets. See also Section~\ref{ssec:conclusion}.  
\subsection{Self-distillation for Semi-supervised Learning}\label{ssec:method-distilation}

The low data regime of the previous section refers to having both few data and few labels. We further propose to investigate the case of semi-supervised learning ~\cite{lee2013pseudo,yalniz2019billion,xie2020self,chen2020big} where a full facial dataset has been collected but only few labels are provided. To this end, we propose a simple self-distillation technique which fully utilizes network pre-training: we use the fine-tuned network to generate in an online manner new labels for training an identically sized student model on unlabeled data. The student is initialized from a pre-trained model trained in a fully unsupervised manner. The self-distillation process is repeated iteratively for T steps, where, at each step, the previously trained model becomes the teacher. Formally, the knowledge transfer is defined as $\argmin_{\theta_{t}} \mathcal{L}( (f(x,\theta_{t-1}), f(x,\theta_{t})))$, where $x$ is the input sample, $\theta_{t-1}$ and $\theta_t$ are the parameters of the teacher and the student, respectively, and $\mathcal{L}$ is the task loss (e.g. pixel-wise $\ell_2$ loss for facial landmark localization).

\section{Ablation Studies}\label{sec:ablation}

In this section, we study and answer key questions related to our approach.

\noindent \textbf{Fine-tuning vs. linear adaptation:} Our results, provided in Table~\ref{tab:results_au}, show that linear adaptation results in significant performance degradation. As our ultimate goal is high accuracy for the end facial task, linear adaptation is not considered for the rest of our experiments.

\noindent \textbf{How much facial data is required?} Unlike supervised, unsupervised pre-training does not require labels and hence it can be applied easily to all types of combinations of facial datasets. Then, a natural question arising is how much data is needed to learn a high-quality representation. To this end, we used 3 datasets of varying size. The first one, comprising $\sim3.3M$ images, is the original VGG-Face dataset (VGG-Face). The second comprises $\sim1M$ images randomly selected from VGGFace2 (VGG-Face-small). The last one, coined as Large-Scale-Face, comprises over 5M images, and is obtained by combining VGG-Face, 300W-LP~\cite{zhu2016face}, IMDb-face~\cite{wang2018devil}, AffectNet~\cite{mollahosseini2017affectnet} and WiderFace~\cite{yang2016wider}. For more details regarding the datasets see Section~\ref{sec:method-datasets}.  We trained 3 models on these datasets and evaluated them for the tasks of facial landmark localization, AU intensity estimation and face recognition. As the results from Table~\ref{tab:results_data_amount} show, VGG-Face vs. VGG-Face-small yields small yet noticeable improvements especially for the case of $2\%$ of labelled data. We did not observe further gains by training on Large-Scale-Face.

\begin{wraptable}[19]{r}{0.5\textwidth}
\caption{Impact of different datasets on the facial representations learned \textit{in an unsupervised manner} for the tasks of facial landmark localization (300W), AU intensity estimation (DISFA) and face recognition (IJB-B).} 
\label{tab:results_data_amount}
\centering
\resizebox{.48\textwidth}{!}{
    \begin{tabular}{lcccc}
    \toprule
    \textbf{\multirow{2}{1.1cm}{\centering Data amount}} & 
    \textbf{\multirow{2}{1.1cm}{\centering Unsup. Data}} & 
    \textbf{300W} & \textbf{DISFA}  & \textbf{IJBB}  
     \\
    \cmidrule(lr){3-5}
    &  &   NME &  ICC & $10^{-4}$  \\
    \midrule
    \multirow{ 4}{*}{100\%}  & VGG-Face-small      & 3.91  & .583 & 0.910  \\
                            & VGG-Face      & 3.85  & \textbf{.598} & \textbf{0.912}  \\
                          & Large-Scale-Face     & \textbf{3.83} & .593 & \textbf{0.912}     \\
                          & Flickr-Face      & 3.86 & .590 & 0.911  \\
    \midrule
    \multirow{ 4}{*}{10\%}  & VGG-Face-small       & 4.37  & .572 & 0.887\\
                            & VGG-Face      & \textbf{4.25}  & .592 & 0.889\\
                          & Large-Scale-Face      & 4.30 & \textbf{.597} & \textbf{0.892}      \\
                           & Flickr-Face       & 4.31 & .581 & 0.887   \\
    \midrule
    \multirow{ 4}{*}{2\%}  & VGG-Face-small       & 5.46 & .550 & 0.729 \\
                            & VGG-Face      & \textbf{4.97}  & .560 & \textbf{0.744}  \\
                          & Large-Scale-Face      & 4.98 & .551 & 0.743    \\
                          & Flickr-Face       & 5.05 & \textbf{.571} & 0.740   \\
    \bottomrule
    \end{tabular}
}
\end{wraptable}

\noindent \textbf{Curated vs. uncurated datasets:}
While the previous section investigated the quantity of data required, it did not explore the question of data quality. While we did not use any labels during the unsupervised pre-training phase, one may argue that all datasets considered are sanitized as they were collected by human annotators with a specific task in mind. In this section, we go beyond sanitized datasets, by experimenting with the newly completely uncurated, in-the-wild, dataset, coined Flickr-Face, introduced in Section~\ref{sec:method-datasets}.

We trained a model on it and evaluated it on the same tasks/datasets of the previous section. Table~\ref{tab:results_data_amount} shows some remarkable results: the resulting model is on par with the one trained on the full VGG-Face dataset (Section~\ref{sec:results} shows that it outperforms all other pre-training methods, too). We believe that this result can pave a whole new way to how practitioners, both in industry and academia, collect and label facial datasets for new tasks and applications.

\noindent \textbf{Pre-training task or data?} In order to fully understand whether the aforementioned gains are coming from the unsupervised task alone, the data, or both, we pre-trained a model on ImageNet dataset using \textit{both} supervised and unsupervised pre-training. Our experiments showed that both models performed similarly (e.g. 4.97\% vs 5.1\% on 300W@2\% of data) and significantly more poorly than models trained on face datasets. We conclude that \textit{both unsupervised pre-training and data} are required for high accuracy.

\noindent \textbf{Effect of unsupervised method:} Herein, we compare the results obtained by changing the unsupervised pre-training method from SwAV to Moco-v2~\cite{he2020momentum}. Table~\ref{tab:results_moco_vs_swav} shows that SwAV largely outperforms Moco-v2, emphasizing the importance of utilizing the most powerful available unsupervised method. Note, that better representation learning as measured on imagenet, doesn't equate with better representation in general~\cite{chen2020exploring}, hence way it's important to validate the performance of different methods for faces too. Furthermore, we evaluated SwAV models using different batch-sizes which is shown to be an important hyper-parameter. We found both models to perform similarly. See SwAV (256) in Table~\ref{tab:results_moco_vs_swav} for the model trained with batch-size 256. With small batch-size training requires less resources, yet we found that it was prolonged by $2\times$. 

\captionsetup{skip=10pt}
\setlength\intextsep{15pt}
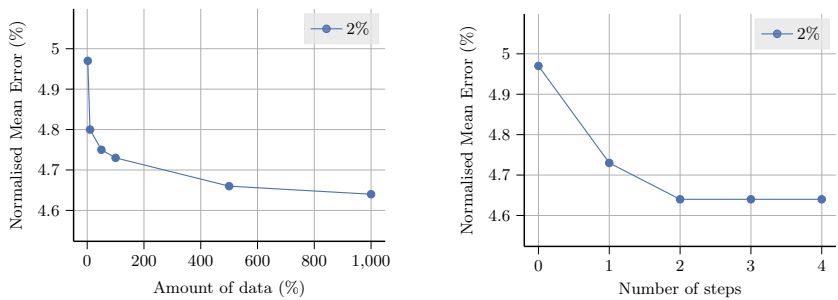
\begin{figure}[ht]
    \centering
    \begin{subfigure}[b]{0.45\textwidth}
            \pgfsetlinewidth{1bp}
             \begin{tikzpicture}[scale=0.7]
\definecolor{color0}{rgb}{0.298039215686275,0.447058823529412,0.690196078431373}

\begin{axis}[
legend cell align={left},
legend style={fill opacity=0.8, draw opacity=1, text opacity=1, draw=white!90!black, fill=white!90!black},
tick align=outside,
tick pos=both,
x grid style={white!69.0196078431373!black},
xlabel={Amount of data (\%)},
xmajorgrids,
xmin=-47.9, xmax=1049.9,
xtick style={color=black},
y grid style={white!69.0196078431373!black},
ylabel={Normalised Mean Error (\%)},
ymajorgrids,
width=7.5cm,height=6cm,
ymin=4.5235, ymax=5.09865,
ytick style={color=black},
ytick={4.6, 4.7, 4.8, 4.9, 5.0},
axis x line*=bottom,
axis y line*=left,
axis line style = thick
]
\addplot [semithick, color0, mark=*, mark size=2, mark options={solid}]
table {%
2 4.96999979019165
10 4.80000019073486
50 4.75
100 4.73000001907349
500 4.65999984741211
1000 4.6399998664856
};
\addlegendentry{2\%}
\end{axis}

\end{tikzpicture}
    \end{subfigure}
    \hfill
    \begin{subfigure}[b]{0.5\textwidth}
            \pgfsetlinewidth{1bp}
             \begin{tikzpicture}[scale=0.7]
\definecolor{color0}{rgb}{0.298039215686275,0.447058823529412,0.690196078431373}

\begin{axis}[
legend cell align={left},
legend style={fill opacity=0.8, draw opacity=1, text opacity=1, draw=white!90!black, fill=white!90!black},
tick align=outside,
tick pos=both,
x grid style={white!69.0196078431373!black},
xlabel={Number of steps},
xmajorgrids,
xmin=-0.2, xmax=4.2,
xtick style={color=black},
y grid style={white!69.0196078431373!black},
ylabel={Normalised Mean Error (\%)},
ymajorgrids,
width=7.5cm,height=6cm,
ymin=4.5235, ymax=5.09865,
ytick style={color=black},
ytick={4.6, 4.7, 4.8, 4.9, 5.0},
axis x line*=bottom,
axis y line*=left,
axis line style = thick
]
\addplot [semithick, color0, mark=*, mark size=2, mark options={solid}]
table {%
0 4.96999979019165
1 4.73000001907349
2 4.6399998664856
3 4.6399998664856
4 4.6399998664856
};
\addlegendentry{2\%}
\end{axis}

\end{tikzpicture}
    \end{subfigure}
    \hfill
    \caption{Self-distillation accuracy for facial landmark vs. (left) amount of unlabeled data ($100\%$ corresponds to 300W), and (right) number of distillation steps. }
    \label{fig:distilation_effect}
\end{figure}

\noindent \textbf{Self-distillation for semi-supervised learning:} Herein, we evaluate the effectiveness of network pre-training on self-distillation (see Section~\ref{ssec:method-distilation}) for the task of semi-supervised facial landmark localization (300W).

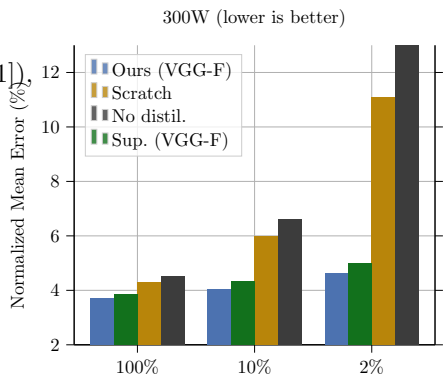
\begin{wrapfigure}[16]{r}{0.5\textwidth}
    \centering
    
            \pgfsetlinewidth{1bp}
             \begin{tikzpicture}[scale=0.7]

\definecolor{color0}{rgb}{0.298039215686275,0.447058823529412,0.690196078431373}
\definecolor{color1}{rgb}{0.72156862745098,0.52156862745098,0.0392156862745098}
\definecolor{color2}{rgb}{0.0705882352941176,0.443137254901961,0.109803921568627}

\begin{axis}[
legend cell align={left},
legend style={fill opacity=0.8, draw opacity=1, text opacity=1, draw=white!80!black},
tick align=outside,
tick pos=both,
x grid style={white!69.0196078431373!black},
xmajorgrids,
xmin=-0.54, xmax=2.54,
xtick style={color=black},
xtick={0,1,2},
xticklabels={100\%,10\%,2\%},
y grid style={white!69.0196078431373!black},
ylabel={Normalized Mean Error (\%)},
ymajorgrids,
ymin=2, ymax=13,
ytick style={color=black},
title={300W (lower is better)},
axis x line*=bottom,
axis line style = thick,
legend pos=north west
]
\draw[draw=none,fill=color0] (axis cs:-0.4,0) rectangle (axis cs:-0.2,3.71);
\addlegendimage{ybar,ybar legend,draw=none,fill=color0};
\addlegendentry{Ours (VGG-F)}

\draw[draw=none,fill=color0] (axis cs:0.6,0) rectangle (axis cs:0.8,4.03);
\draw[draw=none,fill=color0] (axis cs:1.6,0) rectangle (axis cs:1.8,4.64);
\draw[draw=none,fill=color1] (axis cs:0,0) rectangle (axis cs:0.2,4.3);
\addlegendimage{ybar,ybar legend,draw=none,fill=color1};
\addlegendentry{Scratch}

\draw[draw=none,fill=color1] (axis cs:1,0) rectangle (axis cs:1.2,6);
\draw[draw=none,fill=color1] (axis cs:2,0) rectangle (axis cs:2.2,11.1);
\draw[draw=none,fill=white!23.5294117647059!black] (axis cs:0.2,0) rectangle (axis cs:0.4,4.5);
\addlegendimage{ybar,ybar legend,draw=none,fill=white!23.5294117647059!black};
\addlegendentry{No distil.}

\draw[draw=none,fill=white!23.5294117647059!black] (axis cs:1.2,0) rectangle (axis cs:1.4,6.61);
\draw[draw=none,fill=white!23.5294117647059!black] (axis cs:2.2,0) rectangle (axis cs:2.4,13.5);
\draw[draw=none,fill=color2] (axis cs:-0.2,0) rectangle (axis cs:0,3.85);
\addlegendimage{ybar,ybar legend,draw=none,fill=color2};
\addlegendentry{Sup. (VGG-F)}

\draw[draw=none,fill=color2] (axis cs:0.8,0) rectangle (axis cs:1,4.35);
\draw[draw=none,fill=color2] (axis cs:1.8,0) rectangle (axis cs:2,5);
\end{axis}

\end{tikzpicture}
    \caption{Effectiveness of network pre-training on self-distillation for the tasks of facial landmark localization.}
    \label{fig:student_init_distil}
\end{wrapfigure}

We compare unsupervised vs. supervised pre-training on VGG-Face as well as training from scratch. These networks are fine-tuned on 300W using $100\%$ and, the most interesting, $10\%$ and $2\%$ of the data. Then, they are used as students for self-distillation. Fig.~\ref{fig:student_init_distil} clearly shows the effectiveness of unsupervised student pre-training. 

\captionsetup{skip=1pt}
\setlength\intextsep{0pt}

Furthermore, a large pool of unlabelled data was formed by 300W, AFLW~\cite{koestinger2011annotated}, WFLW~\cite{wu2018look} and COFW~\cite{burgos2013robust,ghiasi2015occlusion}), and then used for self-distillation. Fig.~\ref{fig:distilation_effect} (left) shows the impact on the accuracy of the final model by adding more and more unlabelled data to the self-distillation process. Clearly, self-distillation based on network pre-training is capable of effectively utilizing a large amount of unlabelled data. Finally, Fig.~\ref{fig:distilation_effect} (right) shows the impact of the number of self-distillation steps on accuracy. 

\noindent \textbf{Other supervised pre-training:} Our best supervised pre-trained network is that based on training CosFace~\cite{wang2018cosface} on VGG-Face. Herein, for completeness, we compare this to supervised pre-training on another task/dataset, namely facial landmark localization. As Table~\ref{tab:results_sup_pretrain}
shows, the supervised pre-trained model on VGG-Face outperforms it by large margin. This is expected due to the massive size of VGG-Face. 

\addtolength{\tabcolsep}{2pt} 
\begin{table}[ht]
\caption{Supervised pre-training applied to different datasets. The models are evaluated for AU intensity estimation on DISFA. }
\label{tab:results_sup_pretrain}
\centering
\resizebox{.6\textwidth}{!}{
    \begin{tabular}{lcccc}
    \toprule
    \multirow{2}{1.1cm}{\textbf{Data amount}} & 
     \multicolumn{3}{c}{\textbf{Pretrain. method}}
     \\
     \cmidrule(lr){2-4}
      & \sImagenet  & \sVGG & {\small Sup. (300W)} \\ 

    \midrule
    100\% & .560 & .575 & .463 \\ 
    \midrule
    10\% & .556   & .560  & .460 \\ 
    \midrule
    1\% & .453 & .542 & .414 \\ 
    \bottomrule
    \end{tabular}
    }
\end{table}
\addtolength{\tabcolsep}{3pt} 

\section{Main Results}\label{sec:results}

In this section, we thoroughly test the generalizability of the universal facial representations by adapting the resulting models to the most important facial analysis tasks. The full training and implementation details for each of this tasks is detailed in the corresponding sub-section. Training code will be  made available.

\noindent \textbf{Data $\&$ label regime:} For all datasets and tasks, we used fine-tuning for network adaptation using 3 data and label regimes: full (100\%), low (10\%) and very low (2\% or less). For all low data scenarios, we randomly sub-sampled a set of annotated images without accounting for the labels (i.e. we don’t attempt to balance the classes).  Once formed, the same subset is used for all subsequent experiments to avoid noise induced by different sets of images. For face recognition, we deviated slightly from the above setting by enforcing that at least 1/4 of the identities are preserved for the very low data regime of 2\%. This is a consequence of the training objective used for face recognition that is sensitive to both the number of identities and samples per identity.

\noindent \textbf{Models compared:} For unsupervised network pre-training, we report the results of two models, one trained on the full VGG-Face and one on Flickr-Face. These models are denoted as Ours (VGG-F) and Ours (Flickr-F). These models are compared with supervised pre-training on ImageNet and VGG-Face (denoted as VGG-F), as well as the model trained from scratch. 

\noindent \textbf{Comparison with SOTA:} Where possible, we also present the results reported by state-of-the-art methods for each task on the few-shot setting. Finally, for each task, and, to put our results into perspective, we report the accuracy of a state-of-the-art method for the given task. We note however, that the results are not directly comparable, due to different networks, losses, training procedure, and even training datasets. 

\begin{table*}[ht]
\caption{Face recognition results in terms of TAR on IJB-B and IJB-C.} 
\label{tab:results_fr}
\centering
\resizebox{.99\textwidth}{!}{
    \begin{tabular}{lcccccc|ccccc}
    \toprule
    \textbf{\multirow{2}{*}{\begin{tabular}{@{}c@{}}Data \\ amount\end{tabular}}} & 
    \textbf{\multirow{2}{*}{\begin{tabular}{@{}c@{}}Pretrain. \\ method\end{tabular}}} & 
    \multicolumn{5}{c}{IJB-B} & \multicolumn{5}{c}{IJB-C}   \\
    \cmidrule(lr){3-7} \cmidrule(lr){8-12}
     & & $10^{-6}$ & $10^{-5}$& $10^{-4}$ & $10^{-3}$&$10^{-2}$ & $10^{-6}$ & $10^{-5}$& $10^{-4}$ & $10^{-3}$&$10^{-2}$ \\
    \midrule
    \multirow{ 3}{*}{100\%} & \scratch & 0.389 & 0.835 & \textbf{0.912} & 0.950 & 0.975 & 0.778 & 0.883 & 0.931 & 0.961 & 0.981  \\
                          & \sImagenet & 0.390 & \textbf{0.843} & \textbf{0.912} & 0.950 & 0.975  & 0.831 & \textbf{0.891} & 0.931 & 0.961 & 0.981 \\
                          & \oursFlickr & 0.406 & 0.834 & 0.911    & \textbf{0.951} & 0.975 & 0.807 & 0.880 & \textbf{0.932} & \textbf{0.962} & \textbf{0.982}    \\
                          & \oursVGG     & \textbf{0.432} & 0.835 & \textbf{0.912} & 0.950 & \textbf{0.976} & \textbf{0.882} & 0.882 & \textbf{0.932} & 0.961 & 0.981 \\
    \midrule
    \multirow{ 3}{*}{10\%} & \scratch & 0.326 & 0.645 & 0.848   & 0.926 & 0.965 & 0.506 & 0.7671 & 0.8840 & 0.940 & 0.721       \\
                          & \sImagenet & 0.320 & 0.653 & 0.858   & 0.926 & 0.966 & 0.503 & 0.779 & 0.891 & 0.941 & 0.973     \\
                          & \oursFlickr & 0.334 & 0.758 & 0.887    & 0.940 & 0.970 & 0.715 & 0.834 & 0.909 & 0.952 & \textbf{0.978}   \\
                          & \oursVGG     & \textbf{0.392}  & \textbf{0.784} & \textbf{0.889}  & \textbf{0.941} & \textbf{0.972} & \textbf{0.733} & \textbf{0.847} & \textbf{0.911} & \textbf{0.953} & 0.977   \\
    \midrule
    \multirow{ 3}{*}{2\%} & \scratch  & 0.086 & 0.479 & 0.672    & 0.800 & 0.909  & 0.400 & 0.570 & 0.706 & 0.829 & 0.922  \\
                          & \sImagenet & 0.264 & 0.553 & 0.694    & 0.820 & 0.915 & \textbf{0.493} & 0.599 & 0.723 & 0.841 & 0.928   \\
                          & \oursFlickr & 0.282 & \textbf{0.558} & 0.740    & 0.870 & 0.944 & 0.486 & \textbf{0.649} & \textbf{0.786} & 0.891 & 0.954    \\
                          & \oursVGG    & \textbf{0.333}  & 0.547 & \textbf{0.744}    & \textbf{0.873} & \textbf{0.948} & 0.455 & 0.637 & \textbf{0.786} & \textbf{0.893} & \textbf{0.956}   \\
    \bottomrule
     \multicolumn{2}{c}{SOTA (from paper)~\cite{deng2019arcface}}  & 0.401  & 0.821 & 0.907    & 0.950 & 0.978 & 0.0.767 & 0.879 & 0.929 & 0.964 & 0.984   \\
    \bottomrule
    \end{tabular}
    }
\end{table*}

\subsection{Face Recognition}

For face recognition, we fine-tuned the models on the VGGFace~\cite{cao2018vggface2} and tested them on the IJB-B~\cite{whitelam2017iarpa} and IJB-C~\cite{maze2018iarpa} datasets. The task specific head $h(.)$ consists of a linear layer. The whole network was optimized using the CosFace loss~\cite{wang2018cosface}. Note that, for this experiment, since training was done on VGGFace~\cite{cao2018vggface2}, the results of supervised pre-training on VGG-Face are omitted (as meaningless). For training details, see supplementary material.

\noindent\textbf{Results} are shown in Table~\ref{tab:results_fr}. Both \textit{Ours (VGG-F)} and \textit{Ours (Flickr-F)} perform similarly and both they outperform the other baselines by large margin for the low (10\%) and very low (2\%) data regimes. For the latter case, the accuracy drops significantly for all cases.

\subsection{Facial Landmark Localization}

\begin{wraptable}[18]{r}{0.45\textwidth}
\caption{Comparison against state-of-the-art in few-shot facial landmark localization.} 
\label{tab:sota_fa_fewshot}
\centering
\resizebox{.45\textwidth}{!}{
    \begin{tabular}{lccc}
    \toprule
    
    \textbf{300W} & 100\% & 10\% & 1.5\% \\
    \midrule
   RCN+~\cite{honari2018improving} & 3.46 & 4.47 & -  \\
   $\text{TS}^{3}$~\cite{dong2019teacher} & 3.49 & 5.03 & -  \\
   3FabRec~\cite{browatzki20203fabrec} & 3.82 & 4.47 & 5.10  \\
   Ours (VGG-F) & \textbf{3.20} & \textbf{3.48} & \textbf{4.13} \\
    \midrule
    \textbf{AFLW} & 100\%  & 10\% &  1\% \\
    \midrule
   RCN+~\cite{honari2018improving} & 1.61 & - & 2.88  \\
   $\text{TS}^{3}$~\cite{dong2019teacher} & - & 2.14 & -  \\
   3FabRec~\cite{browatzki20203fabrec} & 1.87 & 2.03 & 2.38  \\
   Ours (VGG-F) & \textbf{1.54} & \textbf{1.70} & \textbf{1.91} \\
    \midrule
    \textbf{WFLW} & 100\%  & 10\% & 0.7\% \\
    \midrule
   SA~\cite{qian2019aggregation} & 4.39 & 7.20 & -  \\
   3FabRec~\cite{browatzki20203fabrec} & 5.62 & 6.73 & 8.39  \\
   Ours (VGG-F) & \textbf{4.57} & \textbf{5.44} & \textbf{7.11} \\
    \bottomrule
    \end{tabular}
    }
\end{wraptable}

We fine-tuned the pre-trained models for facial landmark localization on 300W~\cite{sagonas2016300}, AFLW-19~\cite{koestinger2011annotated}, WFLW~\cite{wu2018look} and COFW-68~\cite{burgos2013robust,ghiasi2015occlusion} reporting results in terms of $\text{NME}_{\text{i-o}}$~\cite{sagonas2016300} or $\text{NME}_{\text{diag}}$~\cite{koestinger2011annotated}. We followed the current best practices based on heatmap regression~\cite{bulat2017far}. In order to accommodate for the pixel-wise nature of the task, the task specific head $h(.)$ is defined as a set of 3 $1\times1$ conv. layers with 256 channels, each interleaved with bilinear upsampling operations for recovering part of the lost resolution. Additional high resolution information is brought up via skip connections and summation from the lower part of the network. Despite the simple and un-optimized architecture we found that the network performs very well, thanks to the strong facial representation learned. All models were trained using a pixel-wise MSE loss.  For full training details, see supp. material.

\noindent\textbf{Results} are shown in Table~\ref{tab:results_fa2}: unsupervised pre-training (both models) outperform the other baselines for all data regimes, especially for the low and very low cases. For the latter case, \textit{Ours (VGG-F)} outperforms \textit{Ours (Flickr-F)} probably because \textit{Ours (VGG-F)} contains a more balanced distribution of facial poses. The best supervised pre-training method is VGG-F showing the importance of pre-training on facial datasets.

Furthermore, Table~\ref{tab:sota_fa_fewshot} shows comparison with few very recent works on few-shot face alignment. Our method scores significantly higher across all data regimes and datasets tested setting a new state-of-the-art despite the straightforward network architecture and the generic nature of our method.

\begin{table}[t]
\caption{Facial landmark localization results on 300W (test set), COFW, WFLW and AFLW in terms of $\text{NME}_{\text{inter-ocular}}$, except for AFLW where $\text{NME}_{\text{diag}}$ is used.} 
\label{tab:results_fa2}
\centering
\resizebox{.7\textwidth}{!}{
\begin{tabular}{lccccc}
\toprule
\textbf{\multirow{2}{*}{ \begin{tabular}{@{}c@{}}Data \\ amount\end{tabular}}} & 
\textbf{\multirow{2}{*}{ \begin{tabular}{@{}c@{}}Pretrain. \\ method\end{tabular}}} & 
\textbf{\multirow{2}{*}{300W}} & \textbf{\multirow{2}{*}{COFW}} & \textbf{\multirow{2}{*}{WFLW}} & \textbf{\multirow{2}{*}{AFLW}} \\
&                         &    &  &    &     \\
\midrule
\multirow{ 5}{*}{100\%} & \scratch  & 4.50 &    4.10      & 5.10    & 1.59    \\
                      & \sImagenet  & 4.16 & 3.63 & 4.80 & 1.59   \\
                      & \sVGG  & 3.97  & 3.51  & 4.70  & 1.58   \\
                      & \oursFlickr  & 3.86  & 3.45 & 4.65  & 1.57   \\
                      & \oursVGG      & \textbf{3.85}  & \textbf{3.32}  & \textbf{4.57}  & \textbf{1.55}   \\
\midrule
\multirow{ 5}{*}{10\%} & \scratch  & 6.61  & 5.63   & 6.82   & 1.84  \\
                      & \sImagenet  & 5.15   & 5.32   & 6.56    & 1.81   \\
                      & \sVGG    & 4.55   & 4.46    & 5.87    & 1.77   \\
                      & \oursFlickr & 4.31  & 4.27  & 5.45  & \textbf{1.73}   \\
                      & \oursVGG      & \textbf{4.25}    & \textbf{3.95}    & \textbf{5.44} &  1.74    \\
\midrule
\multirow{ 5}{*}{2\%} & \scratch  & 13.52    & 14.7 &   10.43 &  2.23     \\
                      & \sImagenet  & 8.04    & 8.05 & 8.99  & 2.09     \\
                      & \sVGG   & 5.45     & 5.55  & 6.94  & 2.00   \\
                      & \oursFlickr  & 5.05  & 5.18  & 6.53 & \textbf{1.86}   \\
                      & \oursVGG      & \textbf{4.97}     & \textbf{4.70}   & \textbf{6.29}   & 1.88    \\
\bottomrule
\multicolumn{2}{c}{\small SOTA (from paper)~\cite{wang2020deep}} & 3.85 & 3.45 & 4.60 & 1.57  \\
\multicolumn{2}{c}{SOTA (from paper)~\cite{kumar2020luvli}}  & - & - & 4.37 & 1.39  \\
\bottomrule
\end{tabular}
}
\end{table}

\subsection{Action Unit (AU) Intensity Estimation}

We fine-tuned and evaluated the pre-trained models for AU intensity estimation on the corresponding partitions of BP4D~\cite{valstar2015fera,zhang2014bp4d} and DISFA~\cite{mavadati2013disfa} datasets. The network head $h(.)$ is implemented using a linear layer. The whole network is trained to regress the intensity value of each AU using an $\ell_2$ loss. We report results in terms of intra-class correlation (ICC)~\cite{shrout1979intraclass}. For training details, see supplementary material.

\noindent\textbf{Results} are shown in Table~\ref{tab:results_au}: unsupervised pre-training (both models) outperform the other baselines for all data regimes. Notably, our models achieve very high accuracy even for the case when $2\%$ of data was used. Supervised pre-training on VGG-F also works well. 

Furthermore, Table~\ref{tab:sota_fera2015_fewshot} shows comparison with very recent works on semi-supervised AU intensity estimation. We note that these methods had access to all training data; only the amount of labels was varied. Our methods, although trained under both very low data and label regimes, outperformed them by a significant margin.
   
\begin{table}[ht!]
\caption{AU intensity estimation results in terms of ICC on BP4D and DISFA.} 
\label{tab:results_au}
\centering
\resizebox{.75\textwidth}{!}{
    \begin{tabular}{lccccc}
    \toprule
    \textbf{\multirow{2}{1.1cm}{\centering Data amount}} & 
    \textbf{\multirow{2}{1.1cm}{\centering Pretrain. method}} & 
    \multicolumn{2}{c}{\textbf{DISFA}} & \multicolumn{2}{c}{\textbf{BP4D}} 
     \\
    \cmidrule(lr){3-4} \cmidrule(lr){5-6} 
    &  &   \textbf{finetune} & \textbf{linear} &   \textbf{finetune} & \textbf{linear}  \\

    \midrule
    \multirow{ 5}{*}{100\%} & \scratch  & .318 & - & .617 & -  \\
                          & \sImagenet  & .560 & .316 & .708 & .587   \\
                          & \sVGG   & .575 & .235 & .700 & .564 \\
                          & \oursFlickr   & .590 & \textbf{.373} & .715 & .599 \\
                          & \oursVGG     & \textbf{.598} & .342 & \textbf{.719} & \textbf{.610} \\
                       
    \midrule
    \multirow{ 5}{*}{10\%} & \scratch  & .313 & -    & .622 & -       \\
                          & \sImagenet  & .556 & .300   & .698 & .573      \\
                          & \sVGG   & .560 & .232   & .692 & .564    \\
                          & \oursFlickr   & .581 & \textbf{.352} & .699 & .603 \\
                          & \oursVGG      & \textbf{.592} & .340   & \textbf{.706} & \textbf{.604}    \\

    \midrule
    \multirow{ 5}{*}{1\%} & \scratch  & .237 & -    & .586 & -     \\
                          & \sImagenet  & .453 & .301    & .689 & .564    \\
                          & \sVGG   & .542 & .187    & .690 & .562   \\
                          & \oursFlickr   & \textbf{.571} & .321 & \textbf{.695} & \textbf{.596} \\
                          & \oursVGG      & .560 & \textbf{.326}    & .694 & .592    \\
    \bottomrule
     \multicolumn{2}{c}{\small SOTA (from paper)~\cite{ntinou2021transfer}} & 0.57 & - & 0.72 & - \\
    \bottomrule
    \end{tabular}
}
\end{table}

\begin{table*}[ht]
\caption{Comparison against state-of-the-art on few-shot Facial AU intensity estimation on the BP4D dataset.} 
\label{tab:sota_fera2015_fewshot}
\centering
\resizebox{.8\textwidth}{!}{
\begin{tabular}{lccccccc}
\toprule
\textbf{\multirow{2}{*}{ Method}} & 
\textbf{\multirow{2}{*}{ \begin{tabular}{@{}c@{}}Data \\ amount\end{tabular}}} & 
\multicolumn{5}{c}{\textbf{AU}} &  \textbf{\multirow{2}{*}{ \begin{tabular}{@{}c@{}}Avg. \end{tabular}}} 
 \\
\cmidrule(lr){3-7}
      &    &   \textbf{6} & \textbf{10} & \textbf{12} & \textbf{14} & \textbf{17}  \\
\midrule
KBSS~\cite{zhang2018weakly}      &    1\%          & .760 & .725 & .840 & .445 & .454 & .645    \\
KJRE~\cite{zhang2019joint}        &     6\%        & .710 & .610 & .870 & .390 & .420 & .600     \\
CLFL~\cite{zhang2019context}       &    1\%         & .766 & .703 & .827 & .411 & \textbf{.600} & .680   \\
SSCFL~\cite{sanchez2020semi}    & 2\%  & .766 & .749 & .857 & .475 & .553 & .680  \\
\midrule
\textbf{Ours}        & \textbf{1\%}               & \textbf{.789} & \textbf{.756} & \textbf{.882} & \textbf{.529} & .578 & \textbf{.707}     \\
\bottomrule
\end{tabular}
}
\end{table*}

\subsection{Emotion Recognition}

We observe similar behaviour on the well-established AffectNet~\cite{mollahosseini2017affectnet} for emotion recognition. For details and results, see supplementary material.

\subsection{3D Face Reconstruction}

We fine-tuned all models on the 300W-LP~\cite{zhu2016face} dataset and tested them on AFLW2000-3D~\cite{zhu2016face}. Our task specific head is implemented with a GCN based on spiral convolutions~\cite{lim2018simple}. The network was  trained to minimise the $\ell_1$ distance between the predicted and the ground truth vertices.

\noindent\textbf{Training details:} Since 300W-LP has a small number of identities, during training we randomly augment the data using the following transformations: scaling($0.85\times-1.15\times$), in-plane rotation ($\pm45^o$), and random 10\% translation w.r.t image width and height. Depending on the setting, we trained the model between 120 and 360 epochs using a learning rate of $0.05$, a weight decay of $10^{-4}$ and SGD with momentum (set to 0.9). All models were trained using 2 GPUs.

\noindent\textbf{Results} are shown in Table~\ref{tab:results_fa3d}: it can be seen that, \textit{for all} data regimes, our unsupervised models outperform the supervised baselines. Supervised pre-training on VGG-F also works well. For more results, see supplementary material.

\begin{table*}[t]
\caption{3D face reconstruction reconstruction in terms of NME (68 points) on AFLW2000-3D.} 
\label{tab:results_fa3d}
\centering
\resizebox{.75\textwidth}{!}{
    \begin{tabular}{lccccc}
    \toprule
    \multirow{3}{0.7cm}{\textbf{Data}} & 
     \multicolumn{5}{c}{\textbf{Pretrain. method}}
     \\
     \cmidrule(lr){2-6}
     & \scratch & \multirow{2}{1.2cm}{\centering Sup. (Imagenet)}  & \multirow{2}{1.4cm}{\centering Sup. (VGG-F)} & \multirow{2}{1.5cm}{\centering Ours (Flickr-F)} & \multirow{2}{1.4cm}{\centering Ours (VGG-F)}  \\
     & & & & &  \\
    \midrule
    100\% & 3.70 & 3.58 & 3.51  & 3.53 & \textbf{3.42}  \\
    \midrule
    10\% & 4.72   & 4.06  & 3.82 & 3.81 & \textbf{3.72} \\
    \midrule
    2\% & 7.11 & 6.15 & 4.42 & 4.50 & \textbf{4.31} \\
    \bottomrule
    \multicolumn{6}{c}{SOTA (from paper)~\cite{cheng2020faster}: 3.39 } 
    \\
    \bottomrule
    \end{tabular}
    }
\end{table*}

\section{Discussion and Conclusions}\label{ssec:conclusion}
Several conclusions can be drawn from our results: Unsupervised pre-training followed by task-specific fine-tuning provides very strong baselines for face analysis. For example, we showed that such generically built baselines outperformed recently proposed methods for few-shot/semi-supervised learning (e.g. for facial landmark localization and AU intensity estimation) some of which are based on quite sophisticated techniques. Moreover, we showed that unsupervised pre-training largely boosts self-distillation. Hence, it might be useful for newly-proposed task-specific methods to consider such a pipeline for both development and evaluation especially when newly-achieved accuracy improvements are to be reported. 

Furthermore, these results can be achieved even by simply training on uncurated facial datasets that can be readily downloaded from image repositories. The excellent results obtained by pre-training on Flickr-Face are particularly encouraging. Note that we could have probably created a better and more balanced dataset in terms of facial pose by running a method for facial pose estimation. 

When new datasets are to be collected, such powerful pre-trained networks can be potentially used for minimizing data collection and label annotation labour. Our results show that many existing datasets (e.g. AFLW, DISFA, BP4D, even AffectNet) seem to have a large amount of redundancy. This is more evident for video datasets (e.g. DISFA, BP4D). 

Note that by no means our results imply or suggest that all face analysis can be solved with small labelled datasets. For example, for face recognition, it was absolutely necessary to fine-tune on the whole VGG-Face in order to get high accuracy.

\clearpage
%
%
\bibliographystyle{splncs04}
\bibliography{egbib}

\appendix

\section{Implementation details}

\subsection{Unsupervised pretraining}
For the unsupervised pretraining, similarly with~\cite{caron2020unsupervised} we trained our model on 64 GPUs using a batch size of 4096 and Synchronized Batch Normalization. The network was trained for 200 epochs using a weight decay of  $10^{-6}$ and learning rate of 4.8 that was decayed toward 0.045 using a Cosine Scheduler~\cite{loshchilov2016sgdr}. During the first 10 epochs the learning rate is increased toward the target value using a linear scheduler. In all experiments, unless otherwise specified, we kept the temperature parameter to 0.1 and the Sinkhorn regularization parameters to 0.05. Each input sample was augmented into 2 views at a resolution of $224\times224$px and 6 at a resolution of $96\times96$px. The model was trained using the LARS~\cite{you2017large} optimizer and was implemented in PyTorch~\cite{NEURIPS2019_9015}. 

\noindent\textbf{Datasets and data preparation:} All images are detected using~\cite{deng2019retinaface} and then cropped based on the produced bounding-box so that the face will take approx. 190px on a $256\times256$px image. Unless otherwise specified all the data used for unsupervised pre-training were processed in the same manner.

\subsection{Downstream task implementation details}

Herein, we present the implementation details for each downstream task used in the main body to evaluate the efficacy of the facial representation learned. We note that in all cases the images were normalized in accordance with the training procedure of the pre-trained backbone model used as initialization.

\subsubsection{Face recognition}

Following the best practices~\cite{wang2018cosface,deng2019arcface}, all images were normalized and aligned using the provided 5 landmarks. During training, the only augmentation applied was random horizontal flipping. Depending on the data regime, the models were trained between 18 and 54 epochs using a batch size of 512 and learning rate of $0.1$. The weight decay was set to $0.0005$ and the models were optimized using SGD with momentum (set to $0.9$). For the cosface loss, the margin was set to $0.35$. All models were trained on 8 GPUs. 

\subsubsection{Facial Landmark Localization}

The facial landmark localization pipeline was implemented following~\cite{bulat2017far,sun2019high}. During training, we applied the following augmentations randomly: rotation (between $\pm30^o$), horizontal flipping, scaling ($0.85\times-1.15\times$) and color jittering. Depending on the data regime, dataset and pretrained model, as detailed in the main body of the work, we trained the models between 60 and 480 epochs using a learning rate of $0.0001$, a batch size of 24, a weight decay of $10^{-5}$ and Adam optimizer~\cite{kingma2014adam} ($\beta_1=0.5, \beta_2=0.99$).  All the models were trained using a pixel-wise $\ell_2$ on a single GPU.

\subsubsection{Action Unit (AU) Intensity Estimation}

For AU intensity estimation, we adopted a similar augmentation strategy with the one used for face alignment, mainly we applied random rotation ($\pm30^o$), random horizontal flipping and scale jittering ($0.85\times-1.15\times$), Gaussian blurring with a kernel size between 5 and 10px and a probability of 0.4 and colour jittering. Depending on the setting, the models were trained between 60 and 320 epochs. The learning rate was typically set to $0.0001$, the weight decay to $0.000005$ and the batch size to 48. The models were optimized using Adam ($\beta_1=0.5, \beta_2=0.99$) and trained on 2 GPUs.

\subsubsection{Emotion recognition}

For valence and arousal estimation, we applied the same augmentation strategies as for AU Intensity Estimation with the exception of Gaussian blurring. Depending on the setting, the models were trained between 60 and 240 epochs using a batch size of 32, a learning rate of $0.1$, weight decay of $10^{-4}$ and Adam optimizer($\beta_1=0.5, \beta_2=0.99$). All models were trained on a single GPU.

\subsubsection{3D Face reconstruction}

Since 300W-LP has a small number of identities, during training we randomly augment the data using the following transformations: scaling($0.85\times-1.15\times$), in-plane rotation ($\pm45^o$), and random 10\% translation w.r.t image width and height. Depending on the setting, we trained the model between 120 and 360 epochs using a learning rate of $0.05$, a weight decay of $10^{-4}$ and SGD with momentum (set to 0.9). All models were trained using 2 GPUs.

\subsection{Data sampling}

For all low data scenarios, we randomly subsampled a set of annotated images without accounting for the labels (\textit{i.e.}  we don't attempt to balance the classes). Once formed, the same subset is used for all subsequent experiments to avoid noise induced by  different sets of images. For face recognition where the loss attempts to minimize the intra-class  while maximising the inter-class distance and its sensitivity to both the number of identities and samples per identity, we deviated slightly from the above setting by enforcing that at least 1/4 of the identities are preserved for the very low data regime of 2\%.

\section{Curated Datasets}

For unsupervised pre-training we explore 3 curated datasets, collected for various facial analysis tasks:  (a) Full VGG-Face ($\sim 3.4M$), (b) Small VGG-Face ($\sim1M$) and (c) Large-Scale-Face ($ > 5.0M$), consisting of VGG-Face2~\cite{cao2018vggface2}, 300W-LP~\cite{zhu2016face}, IMDb-face~\cite{wang2018devil}, AffectNet~\cite{mollahosseini2017affectnet} and WiderFace~\cite{yang2016wider}. During unsupervised pre-training we drop all labels using only the facial images. See supplementary material for more details.

a) \textit{Full VGG-Face} denotes the entirety of the VGG-Face2 dataset~\cite{cao2018vggface2}, consisting of $\sim3.4M$ facial images of 9131 identities, with an average of 362.6 images for each subject. Images are downloaded from Google Image Search and have large variations in pose, age, illumination, ethnicity and profession, although they typically depict celebrities.

b) \textit{Small VGG-Face} is a randomly sampled subset of 1M images selected from VGG-Face2.

c) \textit{Large-Scale-Face} is constructed by combining the facial images from VGG-Face2~\cite{cao2018vggface2}, 300W-LP~\cite{zhu2016face}, IMDb-face~\cite{wang2018devil}, AffectNet~\cite{mollahosseini2017affectnet} and WiderFace~\cite{yang2016wider}. Therefore, the dataset combines a set of datasets originally collected for facial recognition, face alignment, emotion recognition and face detection:

\textit{300W-LP}~\cite{zhu2016face} is a face alignment dataset constructed by warping into large poses, from $-90^o$ to $90^o$, the $\sim4000$ near-frontal images from the 300W~\cite{sagonas2013300} dataset.
\textit{IMDb-face}~\cite{wang2018devil} is a large-scale noise-controlled dataset for face recognition, originally containing 1.7M faces with 59,000 identities which were manually cleaned by the authors from 2.0M raw images. All images were obtained by downloading data from the IMDb website.
\textit{AffectNet}~\cite{mollahosseini2017affectnet} is a \textit{in-the-wild} facial expression dataset consisting of more than 1M images collected by queering results from the internet using 1250 emotion related keywords. Out of this, 440,000 images were manually annotated with 7 discrete facial expressions and the intensity of valence and arousal.
\textit{WiderFace}~\cite{yang2016wider} is a face detection benchmarking dataset consisting of 393,703 faces sourced from 32,203 images. The faces exhibit a high degree of variability in terms of scale, pose and occlusion.

\section{Uncurated Flick-Face dataset}

\begin{figure}
    \centering
    \hspace{-0.3cm}
        \begin{tikzpicture}[line join=bevel]
            \useasboundingbox (0,0) rectangle (7.5,3);
            \scope[transform canvas={scale=.75}]
            \pgfsetlinewidth{1bp}
             \input{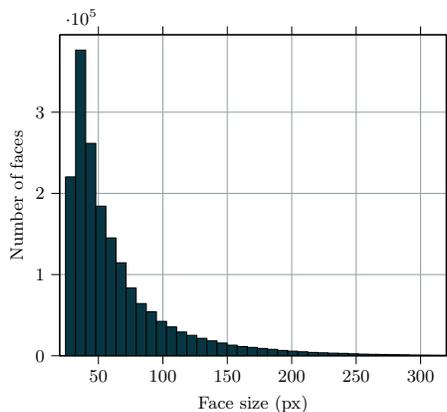}
            \endscope
        \end{tikzpicture}
        \vspace{-1.2cm}
    \caption{Distribution on face sizes in the uncurated Flickr-Face dataset.}
    \label{fig:face_sizes}
\end{figure}

Herein we provide additional details regarding the collected uncurated, in-the-wild, Flickr-Face dataset. The dataset was constructed by downloading a set of images from Flickr. The facial images were then automatically localized and cropped using a face detector~\cite{deng2019retinaface}. In order to increase the likelihood of finding a face in the image we downloaded images that have one of the following 100 tags: \textit{human, people, person, face, fashion, portrait, emotion, expression, affect, happy, sad, anger, angry, smile, laugh, joy, surprise, disgust, confused, fear, horror, adult, lady, ladies, beauty, gentleman, gentlemen, man, men, woman, women, baby, infant, toddler, kid, child, children, senior, father, mother, dad, mom, elderly, grandfather, grandmother, grandpa, grandma, grandparent, ancestor, 40s, 50s, 60s, 70s, 80s, 90s, couple, family, brother, sister, sibling, cousin, wedding, marriage, funeral, party, formal, boy, girl, teen, teenager, youth, friend, classmate, group photo, team, gathering, teacher, professor, lecturer, coach, tutor, worker, boss, celebrity, sport, self, selfie, photoshoot, concert, gigs, band, dance, marathon, passenger, army, soldier, marching, military, protest, crowds}. In total we collected 1.793.119 facial images with a bounding box size that follows the distribution shown in Fig.~\ref{fig:face_sizes}. We release the code used to download the images from Flickr thus allowing reproducing the dataset.

\section{Additional results}

Herein, we report results for AU intensity estimation and emotion recognition (see Section~\ref{ssec:emotion-recognition} and Tables~\ref{tab:sota_disfa_fewshot},~\ref{tab:sota_fera2015_fewshot_full} and~\ref{tab:results_emo}).

\subsection{Emotion Recognition}\label{ssec:emotion-recognition}

We fine-tuned the models for valence and arousal estimation on the well-established AffectNet~\cite{mollahosseini2017affectnet}. We report results in terms of RMSE and CCC~\cite{ringeval2015avec}, SAGR and PCC. The task specific head $h(.)$ is a linear layer that regresses the valence and arousal values and also predicts the basic emotion classes. The network was trained to jointly minimise the RMSE and CCC losses for valence and arousal, and the cross-entropy loss for classification.

\noindent\textbf{Results} are shown in Table~\ref{tab:results_emo} again, \textit{for all} data regimes, our unsupervised models outperform the supervised baselines.

\subsection{Additional 3D Face Reconstruction results}

\begin{figure*}
    \centering
    \begin{subfigure}[b]{0.3\textwidth}
    \includegraphics[width=4.0cm]{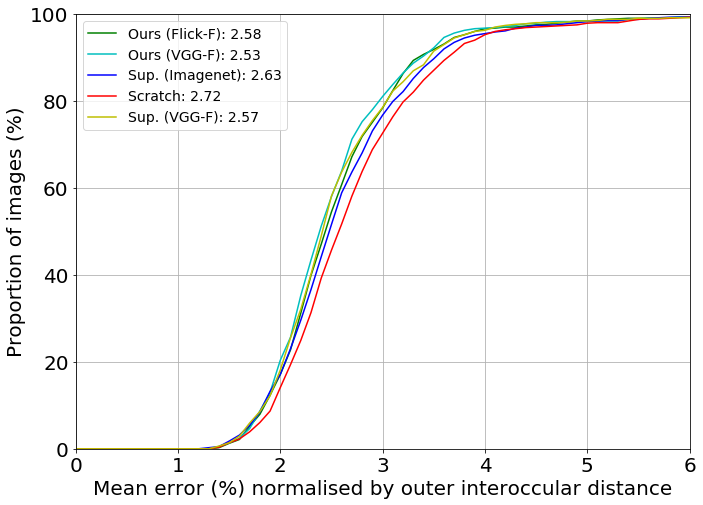}
    \caption{Trained on 100\% of the data.}
    \end{subfigure}
    \hfill
    \begin{subfigure}[b]{0.3\textwidth}
    \includegraphics[width=4.0cm]{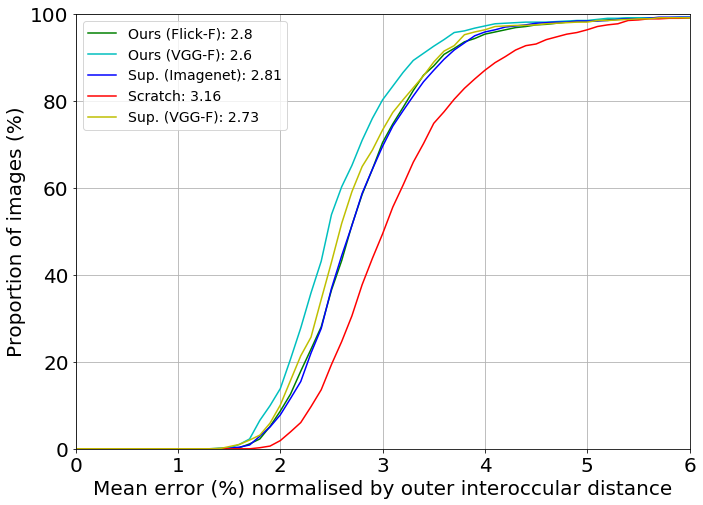}
    \caption{Trained on 10\% of the data.}
    \end{subfigure}
    \hfill
    \begin{subfigure}[b]{0.3\textwidth}
    \includegraphics[width=4.0cm]{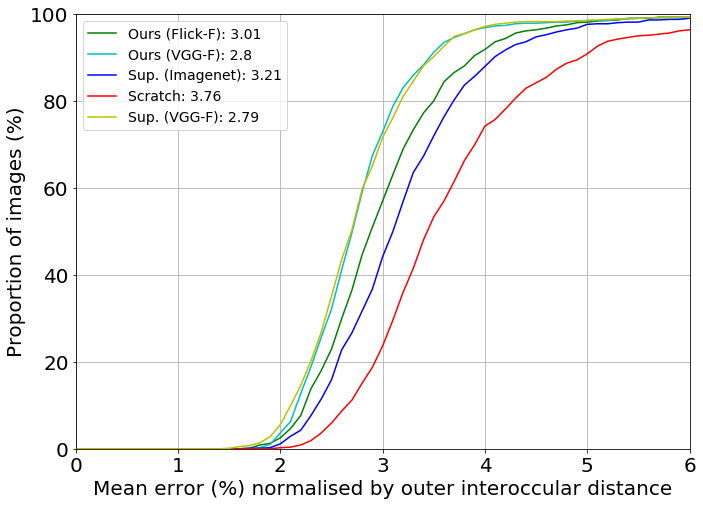}
    \caption{Trained on 2\% of the data.}
    \end{subfigure}
    \caption{Cumulative 3D reconstruction error curves on the Florence~\cite{bagdanov2011florence} dataset for 3 different supervised data regimes: (a) using 100\%, (b) 10\% and (c) 2\%. All models were trained on the 300W-LP dataset as detailed in the main body.}
    \label{fig:florence}
\end{figure*}

Furthermore, in Fig.~\ref{fig:florence} we report results on the Florence dataset for the task of 3D face reconstruction.

\begin{table*}[ht]
\caption{Comparison against state-of-the-art on few-shot Facial AU intensity estimation on the DISFA dataset.} 
\label{tab:sota_disfa_fewshot}
\centering
\resizebox{1.0\textwidth}{!}{
\begin{tabular}{lcccccccccccccc}
\toprule
\textbf{\multirow{2}{*}{ Method}} & 
\textbf{\multirow{2}{*}{ \begin{tabular}{@{}c@{}}Data \\ amount\end{tabular}}} & 
\multicolumn{12}{c}{\textbf{AU}} &  \textbf{\multirow{2}{*}{ \begin{tabular}{@{}c@{}}Avg. \end{tabular}}} 
 \\
\cmidrule(lr){3-14}
      &    &   \textbf{1} & \textbf{2} & \textbf{4} & \textbf{5} & \textbf{6} & \textbf{9} & \textbf{12} & \textbf{15}& \textbf{17} & \textbf{20} & \textbf{25} & \textbf{26} \\
\midrule
KBSS~\cite{zhang2018weakly}      &  1\%            & .136 & .116 & .480 & .169 & .433 & .353 & .710 & .154 & .248 & .085 & .778 & .536 & .350   \\
KJRE~\cite{zhang2019joint}        &     6\%        & .270 & .350 & .250 & .330 & .510 & .310 & .670 & .140 & .170 & .200 & .740 & .250 & .350     \\
CLFL~\cite{zhang2019context}       &    1\%         & .263 & .194 & .459 & .354 & .516 & .356 & .707 & .183 & .340 & \textbf{.206} & .811 & .510  & .408  \\
SSCFL~\cite{sanchez2020semi}    & 2\%  & .327 & .328 & .645 & .024 & \textbf{.601} & .335 & .783 & .181 & .243 & .078 &.882 & .578 & .413  \\
\midrule
Ours        & 1\%               & \textbf{.636} & \textbf{.667} & \textbf{.754} & \textbf{.367} & .549 & \textbf{.535} & \textbf{.820} & \textbf{.313} & \textbf{.541} & .199 & \textbf{.928} & \textbf{.608} & \textbf{.574}    \\
\bottomrule
\end{tabular}
}
\end{table*}

\begin{table*}[ht]
\caption{Comparison against state-of-the-art on few-shot Facial AU intensity estimation on the BU4D dataset.} 
\label{tab:sota_fera2015_fewshot_full}
\centering
\begin{tabular}{lccccccc}
\toprule
\textbf{\multirow{2}{*}{ Method}} & 
\textbf{\multirow{2}{*}{ \begin{tabular}{@{}c@{}}Data \\ amount\end{tabular}}} & 
\multicolumn{5}{c}{\textbf{AU}} &  \textbf{\multirow{2}{*}{ \begin{tabular}{@{}c@{}}Avg. \end{tabular}}} 
 \\
\cmidrule(lr){3-7}
      &    &   \textbf{6} & \textbf{10} & \textbf{12} & \textbf{14} & \textbf{17}  \\
\midrule
KBSS~\cite{zhang2018weakly}      &    1\%          & .760 & .725 & .840 & .445 & .454 & .645    \\
KJRE~\cite{zhang2019joint}        &     6\%        & .710 & .610 & .870 & .390 & .420 & .600     \\
CLFL~\cite{zhang2019context}       &    1\%         & .766 & .703 & .827 & .411 & \textbf{.600} & .680   \\
SSCFL~\cite{sanchez2020semi}    & 2\%  & .766 & .749 & .857 & .475 & .553 & .680  \\
\midrule
\textbf{Ours}        & \textbf{1\%}               & \textbf{.789} & \textbf{.756} & \textbf{.882} & \textbf{.529} & .578 & \textbf{.707}     \\
\bottomrule
\end{tabular}
\end{table*}

\begin{table*}[ht]
\caption{Results on the emotion recogntion task on the AffectNet dataset.} 
\label{tab:results_emo}
\centering
\resizebox{1.0\textwidth}{!}{
\begin{tabular}{lccccccccccc}
\toprule
\textbf{\multirow{2}{*}{ \begin{tabular}{@{}c@{}}Data \\ amount\end{tabular}}} & 
\textbf{\multirow{2}{*}{ \begin{tabular}{@{}c@{}}Init. \\ method\end{tabular}}} & 
\textbf{\multirow{2}{*}{Acc.}} &
    \multicolumn{4}{c}{\textbf{Valence}} & \multicolumn{4}{c}{\textbf{Arousal}} 
 \\
\cmidrule(lr){4-7} \cmidrule(lr){8-11} 
&                         &    & \textbf{RMSE} &   \textbf{SAGR} & \textbf{PCC} &   \textbf{CCC} & \textbf{RMSE} &  \textbf{SAGR} & \textbf{PCC} & \textbf{CCC} \\
\midrule
\multirow{ 3}{*}{100\%} & random  & 0.590 & 0.370    & 0.790 & 0.696  & 0.695 & 0.339  & 0.781 & 0.613 & 0.611  \\
                      & imagenet  & 0.592 & 0.360    & 0.789 & 0.705  & 0.705   & \textbf{0.327} & 0.792 & 0.624 & 0.620 \\
                      & vggface   & 0.601 & 0.369    & \textbf{0.798} & 0.707  & 0.706 & 0.330  & \textbf{0.796} & 0.625 & 0.624  \\
                      & ours      & \textbf{0.602} & \textbf{0.356}    & 0.793 & \textbf{0.711}  & \textbf{0.710}  & 0.328 & 0.793 & \textbf{0.634} & \textbf{0.629}  \\
\midrule
\multirow{ 3}{*}{10\%} & random  & 0.493 & 0.402    & 0.752 & 0.626  & 0.625 & 0.366  & 0.753 & 0.536 & 0.536   \\
                      & imagenet  & 0.548 & 0.383    & \textbf{0.784} & 0.655  & 0.654 & 0.351  & 0.767 & 0.569 & 0.566   \\
                      & vggface   & 0.529 & 0.401    & 0.755 & 0.636  & 0.634 & 0.372  & 0.750 & 0.532 & 0.526  \\
                      & ours      & \textbf{0.562} & \textbf{0.382}    & 0.780 & \textbf{0.678}  & \textbf{0.678} & \textbf{0.344}  & \textbf{0.803} & \textbf{0.600} & \textbf{0.599}   \\
\midrule
\multirow{ 3}{*}{2\%} & random  & 0.419 & 0.453    & 0.727 & 0.515  & 0.515 & 0.400  & 0.747 & 0.423 & 0.422    \\
                      & imagenet  & 0.479 & 0.411    & 0.740 & 0.562  & 0.557 & 0.362  & 0.769 & 0.465 & 0.456   \\
                      & vggface   & \textbf{0.511} & 0.416    & 0.\textbf{778} & 0.610  & \textbf{0.607} & 0.384  & 0.768 & 0.485 & \textbf{0.485} \\
                      & ours      & 0.495 & \textbf{0.370}    & 0.763 & \textbf{0.620}  & 0.593 & \textbf{0.338}  & \textbf{0.794} & \textbf{0.500} & 0.471   \\
\bottomrule
\end{tabular}
}
\end{table*}

\end{document}